\documentclass[10pt,twocolumn,letterpaper]{article}

\usepackage[pagenumbers]{cvpr}

\IfFileExists{axessibility.sty}{\usepackage[accsupp]{axessibility}}{}

\usepackage{tikz}
\usepackage{pgfplots}
\usetikzlibrary{arrows.meta,fit,positioning}
\pgfplotsset{compat=1.18}

\usepackage{microtype}

\definecolor{cvprblue}{rgb}{0.21,0.49,0.74}
\usepackage[breaklinks,colorlinks,allcolors=cvprblue]{hyperref}
\hypersetup{
  pdftitle={Seeing the Evidence, Missing the Answer: Tool-Guided Vision-Language Models on Visual Illusions},
  pdfauthor={Xuesong Wang and Harry Wang}
}

\def\paperID{DataCV-28}
\def\confName{CVPR}
\def\confYear{2026}

\title{Seeing the Evidence, Missing the Answer:\\
Tool-Guided Vision-Language Models on Visual Illusions}

\author{
Xuesong Wang\\
Wayne State University\\
{\tt\small xswang@wayne.edu}
\and
Harry Wang\\
University of Michigan\\
{\tt\small whrui@umich.edu}
}

\begin{document}
\maketitle
\begingroup
\renewcommand{\thefootnote}{}
\footnotetext{This is the author's accepted manuscript of a paper accepted to the 2026 IEEE/CVF Conference on Computer Vision and Pattern Recognition Workshops (CVPRW). \copyright~2026 IEEE. Personal use of this material is permitted. Permission from IEEE must be obtained for all other uses, in any current or future media, including reprinting or republishing this material for advertising or promotional purposes, creating new collective works, for resale or redistribution to servers or lists, or reuse of any copyrighted component of this work in other works.}
\endgroup

\begin{abstract}
Vision-language models (VLMs) exhibit a systematic bias when confronted with
classic optical illusions: they overwhelmingly predict the illusion as
``real'' regardless of whether the image has been counterfactually modified.
We present a tool-guided inference framework for the DataCV 2026 Challenge
(Tasks I and II) that addresses this failure mode without any model training.
An off-the-shelf vision-language model is given access to a small set of generic image
manipulation tools---line drawing, region cropping, side-by-side comparison,
and channel isolation---together with an
illusion-type-routing system prompt that prescribes which tools to invoke for
each perceptual question category.
Critically, every tool call produces a \emph{new, immutable image resource}
appended to a persistent registry, so the model can reference and compose any
prior annotated view throughout its reasoning chain.
Rather than hard-coding illusion-specific modules, this generic-tool-plus-routing
design yields strong cross-structural generalization: performance remained
consistent from the validation set to a test set containing structurally
unfamiliar illusion variants (e.g., Mach Bands rotated from vertical to
horizontal stacking).
We further report three empirical observations that we believe warrant
additional investigation: (i) a strong positive-detection bias likely
rooted in imbalanced illusion training data, (ii) a striking dissociation
between pixel-accurate spatial reasoning and logical inference over
self-generated annotations, and (iii) pronounced sensitivity to image
compression artifacts that compounds false positives.
\end{abstract}

\section{Introduction}
\label{sec:intro}

Recent vision-language models (VLMs) achieve impressive performance across
standard visual benchmarks, yet they remain poorly calibrated when the input
violates familiar statistical regularities.  Classic optical illusions---
Ebbinghaus circles, M\"uller-Lyer arrows, Ponzo converging rails, Poggendorff
diagonals, simultaneous-contrast color patches---are canonical examples of
such violations.  In these settings, models consistently \emph{predict the
illusion} even when the image has been counterfactually altered so that the
stimulus is objectively unambiguous~\cite{rahmanzadehgervi2024illusion}.
We argue that this failure cannot be attributed solely to weak visual
representations; rather, it reflects a \emph{training-data imbalance}:
essentially all illusion images seen during pre-training are genuine illusions,
so the model learns that ``illusion-looking'' images always exhibit the
corresponding perceptual effect.  When asked about a structurally similar image
that has been surgically modified to remove the effect, the prior overwhelms
direct visual evidence, producing near-systematically positive detections.

The DataCV 2026 Challenge operationalizes exactly
this problem. Task~I builds on the VI-Probe benchmark of classic visual
illusions~\cite{sun2026viprobe} and presents binary (Yes/No) questions about
a classic illusion image, where some of the images have been modified so that
the apparent illusion does not actually hold. Task~II builds on the
VIA-Bench benchmark of visual illusions and anomalies~\cite{hou2026viabench}
and extends the setup to multiple-choice questions about real-world visual
anomalies and impossible scenes.  Both tracks are \emph{inference-only}:
participants may not fine-tune
or update model weights in any form; only prompting, routing, and tool use are
permitted. In particular, Task~I forbids deterministic measurement-based
pipelines, so our approach restricts itself to non-quantitative visual aids
such as annotation and zoom/crop rather than explicit geometric or pixel-level
decision rules.

Our submission proposes an \textbf{image annotation agent} built around
Gemini Flash-family models~\cite{gemini3flash_modelcard_2025,gemini31flashlite_modelcard_2026}.
Inspired by Visual Sketchpad~\cite{hu2024visualsketchpad} and
ViperGPT~\cite{suris2023vipergpt}, the agent iteratively calls a small library
of generic image manipulation tools.  A shared core tool set---draw a
reference line, crop a region, and place crops side by side---is used in both
tasks, while Task~II additionally exposes analysis-oriented operators such as
channel extraction, blur, and grid overlay to support broader real-world
anomaly categories.
These tools generate auxiliary visual evidence before the model commits to an answer.
Each tool invocation produces a \emph{new immutable image resource} appended
to a persistent in-context registry; the model may revisit any prior annotated
state to compose richer analyses.
Rather than constructing dedicated illusion-type modules (cf.\ VisProg~\cite{gupta2023visprog}),
we embed per-category tool-use strategies directly in the system prompt,
yielding a single generic agent that generalizes across question types.

Our main contributions are:
\begin{itemize}
  \item A tool-guided visual annotation framework for open-ended perceptual VQA
        that operates without any model training.
  \item An \emph{image resource versioning} system that accumulates
        annotated views as immutable named assets, enabling multi-step
        visual reasoning chains with full history access.
  \item A \emph{prompt-routing} design that maps each perceptual question
        category to a recommended tool-call strategy inside the system
        prompt, as an alternative to hard-coded module libraries.
  \item Empirical observations on three systematic VLM failure modes in
        illusion detection that point toward concrete future research
        directions.
\end{itemize}

\section{Related Work}
\label{sec:related}

\paragraph{Tool-guided visual reasoning.}
ViperGPT~\cite{suris2023vipergpt} pioneered the idea of having an LLM
synthesize executable Python programs that call a fixed visual API
(detection, VQA, crop, etc.) to answer compositional visual questions.
VisProg~\cite{gupta2023visprog} extends this with a structured program
synthesis prompt and a larger module library; crucially, every new question
type requires adding a new module.  Our approach shares the compositional
spirit but inverts the design: rather than generating code or maintaining
a module library, the agent calls a small set of general-purpose image
operators guided by natural-language strategy descriptions embedded in
the system prompt.  This removes the module-authoring burden and, as we
show, generalizes better across structurally unfamiliar illusion variants.

\paragraph{Visual chain-of-thought.}
Visual Sketchpad~\cite{hu2024visualsketchpad} is the closest precursor to
our work.  It equips a VLM with the ability to draw geometric annotations
(lines, circles, bounding boxes) on images as visual intermediate steps,
effectively extending chain-of-thought prompting~\cite{wei2022cot} to the
visual modality.  We adopt this core insight and extend it with
(i) an immutable resource versioning system so that \emph{every}
annotated image is permanently retrievable, (ii) a broader and more
diverse tool set tailored for perceptual illusion analysis, and
(iii) a systematic question-type routing strategy in the system prompt.
MM-ReAct~\cite{yang2023mmreact} applies the ReAct framework~\cite{yao2023react}
to multimodal settings with external tool calls; our agentic loop is
architecturally similar but specialized for the illusion-verification task.

\paragraph{Set-of-Mark prompting.}
SoM~\cite{yang2023setofmark} overlays numeric labels on image regions to
improve VLM grounding; this is complementary to our approach, and
similarly demonstrates that augmenting the image view available to the
model at inference time is a productive strategy.

\paragraph{VLMs and visual illusions.}
Rahmanzadehgervi~\etal~\cite{rahmanzadehgervi2024illusion} conduct a
broad evaluation of VLMs on geometric and perceptual tasks and show
that even state-of-the-art models fail at tasks trivially solved by
humans.  Our work takes a complementary, prescriptive stance: given
that the base model fails, how can tool-guided inference recover?

\section{Method}
\label{sec:method}

\subsection{System Overview}

\begin{figure}[t]
  \centering
  \resizebox{\columnwidth}{!}{%
  \begin{tikzpicture}[
    font=\footnotesize,
    >=Latex,
    node distance=4mm,
    stage/.style={
      draw,
      rounded corners=2pt,
      align=center,
      fill=gray!10,
      text width=0.82\columnwidth,
      inner sep=4pt
    },
    panel/.style={
      draw,
      rounded corners=2pt,
      align=left,
      fill=blue!5,
      text width=0.82\columnwidth,
      inner sep=4pt
    },
    flow/.style={->, thick}
  ]
    \node[stage] (input) {%
      \textbf{Input sample}\\
      original image + question prompt
    };

    \node[stage, below=of input] (routing) {%
      \textbf{Prompt-based routing}\\
      classify the question type and choose a tool-use strategy
    };

    \node[panel, below=6mm of routing] (registry) {%
      \textbf{Image registry}\\
      \texttt{original}\\
      \texttt{img\_001}: line overlay\\
      \texttt{img\_002}: crop\\
      \texttt{img\_003}: side-by-side compare\\
      \texttt{img\_\{k\}}: later hypotheses
    };

    \node[panel, below=4mm of registry] (tools) {%
      \textbf{Tool loop}\\
      shared core: draw / crop / compare\\
      Task~II adds: grid / channels / color / blur\\[1mm]
      inspect outputs and decide the next action
    };

    \node[
      draw,
      rounded corners=3pt,
      dashed,
      inner sep=4pt,
      fit=(registry)(tools),
      label={[font=\footnotesize]above:\textbf{Immutable reasoning state}}
    ] (loopbox) {};

    \node[stage, below=7mm of loopbox, fill=green!8] (output) {%
      \textbf{Structured output}\\
      Task~I: binary output \qquad Task~II: four-way output\\
      shared core interface, task-specific tools, prompt, and schema
    };

    \draw[flow] (input) -- (routing);
    \draw[flow] (routing) -- (loopbox);
    \draw[flow] (loopbox) -- (output);
    \draw[flow] (registry.east) to[out=0, in=0, looseness=1.35]
      node[font=\scriptsize, right, align=center] {agent selects\\existing resource}
      (tools.east);
    \draw[flow] (tools.west) to[out=180, in=180, looseness=1.35]
      node[font=\scriptsize, left, align=center] {tool creates\\new resource}
      (registry.west);
  \end{tikzpicture}
  }
  \caption{%
    Pipeline overview.  The agent receives an image and question, routes the
    sample to a question-type-specific strategy, and iteratively calls a shared
    core tool set, with Task~II additionally exposing analysis-specific tools.
    Every tool call creates a new immutable image resource, and the full
    registry remains available throughout the reasoning chain before the agent
    emits a structured Task~I or Task~II answer.
  }
  \label{fig:pipeline}
\end{figure}
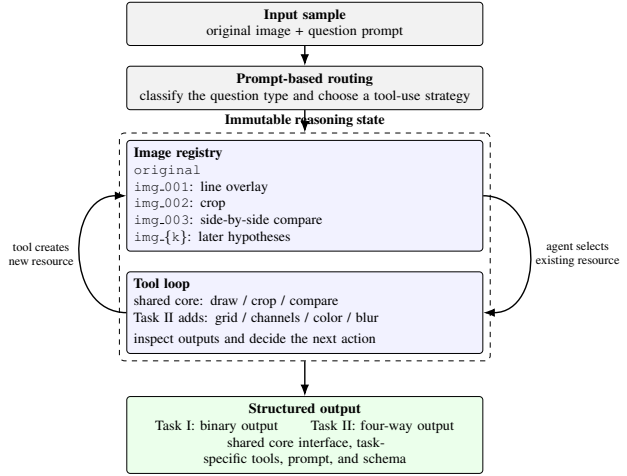

Figure~\ref{fig:pipeline} illustrates the pipeline.
For each sample, an agent is initialized with
the original image registered under the fixed resource ID \texttt{"original"},
the pixel dimensions of the image, and the question text.
The agent then executes a ReAct-style~\cite{yao2023react} loop: it inspects
the current image registry, calls one or more tools, receives the resulting
annotated image(s), and continues until it has sufficient visual evidence
to output a structured answer.
The backbone is a Gemini Flash-family VLM~\cite{gemini3flash_modelcard_2025,gemini31flashlite_modelcard_2026},
used with temperature $0$ and a request limit of 10 tool-call rounds per
sample.  All inference is
zero-shot; no few-shot examples are injected for either task.

\subsection{Image Resource Versioning}
\label{sec:versioning}

A central design decision is that \textbf{no tool ever modifies an existing
image in place}.
Each tool call allocates a new resource---assigned a sequential identifier
\texttt{img\_001}, \texttt{img\_002}, etc.---and the source image is
copied before any drawing or cropping is applied.
The full registry (\texttt{original}, \texttt{img\_001}, \ldots,
\texttt{img\_\{k\}}) persists in context for the lifetime of the sample.

This versioning scheme confers two practical benefits.
First, the model can freely \emph{branch} its annotation path: drawing a
reference line on \texttt{img\_001} and simultaneously cropping a region on
\texttt{img\_002}, then composing them in a later step.
Second, every intermediate visual hypothesis is preserved as a first-class
object; the model can cite, revisit, or re-crop any prior intermediate output,
enabling multi-step reasoning chains that would be lost if a single
mutable image were maintained.
This is analogous to immutable value semantics in functional programming,
applied to the visual modality.

\subsection{Generic Tools with Prompt-Based Routing}
\label{sec:tools}

Following ViperGPT~\cite{suris2023vipergpt} and
Visual Sketchpad~\cite{hu2024visualsketchpad}, our framework
exposes image manipulation as callable tools.
However, whereas VisProg~\cite{gupta2023visprog} maintains a dedicated
\emph{module} for each visual operation type (and requires authors to extend
the library for new question types), we provide a compact set of
\emph{general-purpose} operators organized as a shared core plus
task-specific extensions:

\begin{itemize}
  \item \textbf{Shared core (Task~I and Task~II):}
        \textbf{draw\_line / draw\_rectangle / draw\_circle} for geometric
        annotation, \textbf{crop} for local zoom-in inspection, and
        \textbf{compare\_crops} for side-by-side comparison of candidate
        regions.  These tools cover the measurement-free geometric and
        appearance checks used in both tasks.
  \item \textbf{Task~II extensions:} \textbf{overlay\_grid /
        extract\_channel / sample\_color / isolate\_color / blur} for
        categorical grouping, channel separation, point color sampling,
        color-family isolation (Ishihara tests), and large-radius blur for
        revealing hidden patterns.
\end{itemize}

The key insight is that \emph{how to compose these primitives} for each
illusion type is encoded not in the tools themselves but in the system
prompt.  The prompt first instructs the agent to classify the incoming
question into a task-specific perceptual category (e.g., size
comparison, color comparison, line straightness, hidden content,
impossible figure) and then provides a step-by-step recommended tool
sequence for that category.

\paragraph{Comparison with VisProg.}
VisProg generates a Python program calling named modules
(e.g., \texttt{CROP}, \texttt{VQA}, \texttt{EVAL}).
New question types require new modules, and the model must learn to emit
correct module names.  Our approach replaces the module library with
free-form tool calls guided by natural-language strategies.  This removes
the need to design new modules when the test distribution shifts, and
avoids the code-execution overhead.  The trade-off is that the routing
logic lives in the system prompt and depends on the model's instruction
following.  We therefore inspect tool-call logs directly in
Section~\ref{sec:experiments}; the observed distributions in
Figure~\ref{fig:tool_usage} show that the agent largely follows the
prescribed category-specific strategies, suggesting that a well-structured
prompt is a viable substitute for a hard-coded module registry.

\subsection{Task-Specific Adaptations}

Task~I (binary classification) and Task~II (four-way MCQ) share the same
immutable image-resource versioning scheme and the same geometric
annotation/cropping core, but they differ in their accessible tool subsets,
system prompts, question-type taxonomies, and output schemas.
Task~I's taxonomy covers seven question categories aligned with the
system prompt (size comparison, color comparison, line length,
line straightness, line alignment, line parallelism, and boundary
detection), and therefore only requires the core
drawing and crop/compare tools.
Task~II's taxonomy is substantially broader, spanning sixteen
system-prompt categories ranging from counting and hidden-content
recovery to impossible figures, forced perspective / scale tricks,
entity realism, spatial relation / support, and physical plausibility /
affordance, so we expose the additional analysis tools listed above only
for that track.
A lightweight \emph{rescue agent} with a reduced request limit handles
samples that exhaust the main agent's token budget, ensuring every
sample receives a valid output.

\section{Experiments}
\label{sec:experiments}

\paragraph{Setup.}
We evaluated on the official validation and test splits of both tasks, using
the challenge data released for Task~I (VI-Probe)~\cite{sun2026viprobe} and
Task~II (VIA-Bench)~\cite{hou2026viabench}. Our archived runs use Gemini
Flash-family models with temperature $0$,
a maximum of $10$ agent rounds.
The default model is \texttt{gemini-3-flash-preview}~\cite{gemini3flash_modelcard_2025};
when provider quota temporarily blocked that model, some archived test repairs
instead used \texttt{gemini-3.1-flash-lite-preview}~\cite{gemini31flashlite_modelcard_2026}.
We therefore describe the system as
using Gemini Flash-family backbones rather than a single Gemini variant.
Submissions were scored by the Codabench evaluation server; accuracy is
reported as overall classification accuracy.

\begin{figure}[t]
  \centering
  \begin{tikzpicture}
    \begin{axis}[
      width=0.92\columnwidth,
      height=3.2cm,
      xbar,
      xmin=0, xmax=80,
      xtick={0,20,40,60,80},
      symbolic y coords={compare\_crops,draw\_line,crop,draw\_rectangle,draw\_circle},
      ytick=data,
      y dir=reverse,
      bar width=3.5pt,
      enlarge y limits=0.18,
      tick label style={font=\scriptsize},
      yticklabel style={font=\ttfamily\tiny},
      title={Task I},
      title style={font=\scriptsize\bfseries},
      xmajorgrids=true,
      grid style={dashed,gray!30},
      axis x line*=bottom,
      axis y line*=left,
      point meta=x,
      nodes near coords={\pgfmathprintnumber[fixed,precision=1]{\pgfplotspointmeta}\%},
      every node near coord/.append style={font=\tiny, anchor=west, xshift=1pt},
    ]
      \addplot[fill=blue!55, draw=blue!70!black] coordinates {
        (71.3,compare\_crops)
        (40.6,draw\_line)
        (30.5,crop)
        (0.6,draw\_rectangle)
        (0.2,draw\_circle)
      };
    \end{axis}
  \end{tikzpicture}

  \begin{tikzpicture}
    \begin{axis}[
      width=0.92\columnwidth,
      height=5.1cm,
      xbar,
      xmin=0, xmax=80,
      xtick={0,20,40,60,80},
      xlabel={Samples using tool (\%)},
      symbolic y coords={crop,overlay\_grid,sample\_color,compare\_crops,isolate\_color,draw\_line,blur,draw\_circle,extract\_channel},
      ytick=data,
      y dir=reverse,
      bar width=3.5pt,
      enlarge y limits=0.12,
      tick label style={font=\scriptsize},
      yticklabel style={font=\ttfamily\tiny},
      title={Task II},
      title style={font=\scriptsize\bfseries},
      xmajorgrids=true,
      grid style={dashed,gray!30},
      axis x line*=bottom,
      axis y line*=left,
      point meta=x,
      nodes near coords={\pgfmathprintnumber[fixed,precision=1]{\pgfplotspointmeta}\%},
      every node near coord/.append style={font=\tiny, anchor=west, xshift=1pt},
    ]
      \addplot[fill=orange!65, draw=orange!80!black] coordinates {
        (58.7,crop)
        (27.7,overlay\_grid)
        (19.6,sample\_color)
        (16.0,compare\_crops)
        (14.6,isolate\_color)
        (5.6,draw\_line)
        (4.6,blur)
        (3.6,draw\_circle)
        (1.1,extract\_channel)
      };
    \end{axis}
  \end{tikzpicture}
  \caption{Observed tool-use in archived test-run snapshots. Percentages
  denote the fraction of samples that invoked each tool at least once.
  Task~I remains concentrated on the shared geometric core, while Task~II
  activates the broader analysis-oriented tool set only when needed.}
  \label{fig:tool_usage}
\end{figure}
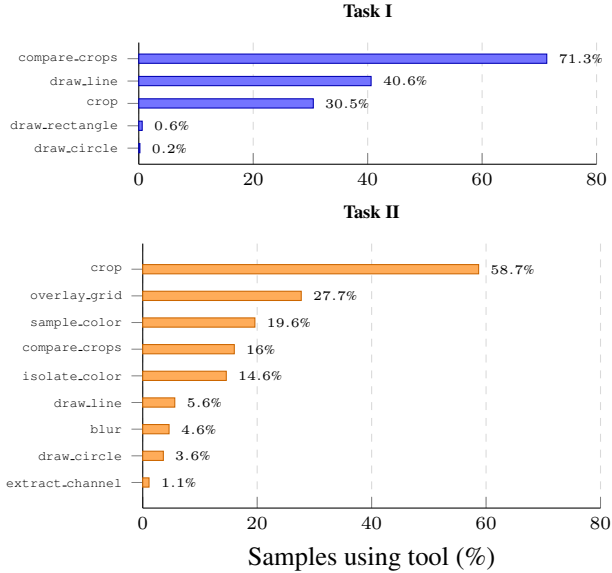

\paragraph{Observed tool-use patterns.}
Because our routing policy is prompt-defined rather than hard-coded, we
inspected the saved tool-call logs from the archived test-run snapshots.
Figure~\ref{fig:tool_usage} shows a clear separation between the two tracks.
Task~I is dominated by the shared geometric core, especially
\texttt{compare\_crops}, \texttt{draw\_line}, and \texttt{crop}, whereas
Task~II selectively activates analysis-oriented operators such as
\texttt{overlay\_grid}, \texttt{sample\_color}, and \texttt{isolate\_color}.
Rare operators such as \texttt{blur} and \texttt{extract\_channel} remain
niche, which is consistent with their prompt role as specialized tools for
hidden-content cases rather than default actions.

\paragraph{Validation--test consistency.}
A key empirical finding is the \emph{stability} of our agent's accuracy
between the validation set and the public test set, despite the test set
containing structurally novel illusion variants not seen during prompt
development.
For example, the validation set's boundary-detection questions feature
vertically stacked color regions (Cornsweet stimulus), while the test
set introduces horizontally stacked variants similar to Mach Bands.
Because the boundary detection strategy in our prompt describes the
\emph{conceptual} procedure (crop around adjacent interfaces, inspect for a
distinct separator line or gap, and check multiple boundaries) rather than
hardcoded pixel
operations, the agent applied the identical strategy successfully to the
horizontally oriented variant without any prompt modification.
This stands in contrast to a niche-tool design (as in
VisProg~\cite{gupta2023visprog}) where a module authored for vertical
boundary detection might silently fail on horizontal orientation.

\section{Observations and Discussion}
\label{sec:discussion}

Beyond task performance, our experiments surfaced three systematic
phenomena that we believe deserve further investigation.

\subsection{Systematic Positive-Detection Bias}
\label{sec:bias}

Across all evaluated models and prompt ablations, VLMs exhibit a
strong tendency to predict \emph{``yes the illusion holds''}
(or the corresponding multiple-choice option) regardless of whether
the image has been modified to break it.
When evaluated on the full Task~I validation set without tool access,
accuracy is well below chance on the negative half (modified images)
while remaining near-perfect on positive samples, indicating a
near-degenerate positive prior.

We hypothesize that the root cause is a \textbf{training-data imbalance}.
The overwhelming majority of illusion images that appear in VLM pre-training
corpora are genuine, unmodified illusions---images captioned ``Ebbinghaus
circles look equal but are different'' or ``these lines appear bent but are
actually straight.''  Counterfactually modified variants that \emph{look
like} an illusion but \emph{are not} one are nearly absent from public data.
Consequently, when the vision encoder produces a representation that
resembles a known illusion template, the language model assigns overwhelming
probability to the positive category.

A more speculative alternative---that the model \emph{inherits} human-like
perceptual vulnerabilities from training on human-generated text---is
unlikely to be the primary driver.  Human visual illusions arise from
specific early-vision mechanisms that process spatial frequency, luminance
concurrency, and relative length in parallel; a transformer operating on
patched embeddings does not obviously replicate these mechanisms.  Moreover,
human illusion susceptibility is \emph{symmetric} (subjects can be fooled in
both directions depending on stimulus design), whereas the model exhibits a
unidirectional over-prediction of positive cases---consistent with a data
bias rather than an inherited perceptual mechanism.

\subsection{Spatial Precision Without Logical Consequence}
\label{sec:precision}

One of the more striking observations from inspecting the agent's tool-use
transcripts is the contrast between the model's \emph{spatial reasoning
accuracy} and its \emph{inferential reliability over its own outputs}.

For line-straightness questions (e.g., Hering illusion), the prescribed
strategy asks the agent to draw a perfectly straight horizontal reference
line at the same $y$-coordinate as the target line and then crop the
region containing both lines for local inspection.
In the vast majority of trials, the model places that reference line at an
appropriate location, and the drawn line is geometrically perfect by
construction.
However, despite producing an annotated image where the target line clearly
curves away from the straight reference, the model often reports that the
lines \emph{are} straight---as if the visual confirmation of its own
correctly placed annotation is overridden by the prior belief that
``Hering-looking lines are straight.''

An analogous pattern appears in Task~II color-comparison questions.
When \texttt{sample\_color} returns two hex codes that differ only in the
least-significant digit (e.g., \texttt{\#8F8F8F} vs.\ \texttt{\#909090}),
the model correctly identifies a numerical difference---yet then concludes
the colors are ``effectively the same,'' thereby predicting the positive
illusion case.  This threshold-setting behavior is not systematically
coded anywhere in our prompt; it appears to emerge from the training
distribution's implicit convention that ``small differences don't count.''
The result is a system that is exquisitely sensitive to pixel-level
differences in annotation placement but surprisingly imprecise in
translating those differences into correct binary conclusions.

\subsection{Image Compression and Noise Sensitivity}
\label{sec:compression}

Many illusion datasets involve synthetic images rendered at low bit depth or
subsequently JPEG-compressed, introducing blocking artifacts and
luminance quantization noise at region boundaries.  Our experiments
indicate that these artifacts interact badly with both the model's
calibration over color-tool outputs and its boundary-detection logic.

For Task~II color-comparison questions, JPEG chroma subsampling can shift a single
region's average color by several quantization steps, causing
\texttt{sample\_color} to return numerically distinct hex codes for
regions that were synthesized with identical pigments.  In contrast to the
cases in Section~\ref{sec:precision}, where the model sometimes dismisses
small real differences as ``effectively the same,'' compression artifacts can
push it in the opposite direction: ratifying a spurious numeric discrepancy as
a true physical difference and predicting a negative outcome (no illusion) on
a positive instance.

For boundary-detection questions, compression can introduce a spurious
gray fringe at color-region interfaces that the model interprets as a
boundary line, producing false positives.
The interaction is compounding: the model is poorly calibrated around small
pixel-level differences (see Section~\ref{sec:precision}), and JPEG
compression reliably creates such differences at every region interface.
Mitigating this likely requires either prompt-level thresholding advice
(``a difference of fewer than $N$ hex levels is not a true boundary'')
or a preprocessing step that normalizes images before tool inspection.

\section{Conclusion}
\label{sec:conclusion}

We presented a tool-guided framework for visual illusion
classification that achieves consistent performance across structurally
varied illusion types without any model training.
Our two principal design choices---immutable image resource versioning and
prompt-embedded routing strategies instead of task-specific module
libraries---proved complementary: versioning enables compositional
multi-step visual reasoning, while prompt routing provides structural
generalization without requiring new module authorship for each novel
question type.

Our qualitative analysis highlights that current VLMs are not simply
``bad at illusions'' in a uniform sense.  They are remarkably precise at
locating objects and annotating geometry, yet they fail to draw correct
conclusions from their own evidence when training-data priors conflict
with visual observation.  This dissociation between perceptual precision
and inferential reliability suggests that improving VLM robustness on
counterfactual stimuli may require targeted data augmentation with
negative illusion instances at least as much as architectural changes.

{
    \small
    \bibliographystyle{ieeenat_fullname}
    \bibliography{main}

@String(CVPR= {IEEE Conf. Comput. Vis. Pattern Recog.})

@String(ICCV= {Int. Conf. Comput. Vis.})

@String(NIPS= {Adv. Neural Inform. Process. Syst.})

@String(ACCV  = {ACCV})

@String(ICLR = {Int. Conf. Learn. Represent.})

@String(CVPR  = {CVPR})

@String(ICCV  = {ICCV})

@String(NIPS  = {NeurIPS})

@String(ICLR  = {ICLR})

@inproceedings{hu2024visualsketchpad,
author = {Hu, Yushi and Shi, Weijia and Fu, Xingyu and Roth, Dan and Ostendorf, Mari and Zettlemoyer, Luke and Smith, Noah A and Krishna, Ranjay},
booktitle = NIPS,
doi = {10.52202/079017-4423},
editor = {A. Globerson and L. Mackey and D. Belgrave and A. Fan and U. Paquet and J. Tomczak and C. Zhang},
pages = {139348--139379},
title = {Visual Sketchpad: Sketching as a Visual Chain of Thought for Multimodal Language Models},
volume = {37},
year = {2024}
}

@inproceedings{suris2023vipergpt,
  title={Vipergpt: Visual inference via python execution for reasoning},
  author={Sur{\'\i}s, D{\'\i}dac and Menon, Sachit and Vondrick, Carl},
  booktitle=ICCV,
  pages={11888--11898},
  year={2023}
}

@inproceedings{gupta2023visprog,
  title={Visual programming: Compositional visual reasoning without training},
  author={Gupta, Tanmay and Kembhavi, Aniruddha},
  booktitle=CVPR,
  pages={14953--14962},
  year={2023}
}

@article{wei2022cot,
title={Chain-of-thought prompting elicits reasoning in large language models},
author={Wei, Jason and Wang, Xuezhi and Schuurmans, Dale and Bosma, Maarten and Xia, Fei and Chi, Ed and Le, Quoc V and Zhou, Denny and others},
journal=NIPS,
volume={35},
pages={24824--24837},
year={2022}
}

@inproceedings{yao2023react,
  title = {{ReAct}: Synergizing Reasoning and Acting in Language Models},
  author = {Yao, Shunyu and Zhao, Jeffrey and Yu, Dian and Du, Nan and Shafran, Izhak and Narasimhan, Karthik and Cao, Yuan},
  booktitle = ICLR,
  year = {2023}
}

@article{yang2023mmreact,
  title={{MM-ReAct}: Prompting {ChatGPT} for Multimodal Reasoning and Action},
  author={Yang, Zhengyuan and Li, Linjie and Wang, Jianfeng and Lin, Kevin and Azarnasab, Ehsan and Ahmed, Faisal and Liu, Zicheng and Liu, Ce and Zeng, Michael and Wang, Lijuan},
  journal={arXiv preprint arXiv:2303.11381},
  year={2023}
}

@article{yang2023setofmark,
  title   = {Set-of-Mark Prompting Unleashes Extraordinary Visual Grounding
             in {GPT-4V}},
  author  = {Yang, Jianwei and Zhang, Hao and Li, Feng and Zou, Xueyan and
             Li, Chunyuan and Gao, Jianfeng},
  journal = {arXiv preprint arXiv:2310.11441},
  year    = {2023}
}

@inproceedings{rahmanzadehgervi2024illusion,
  title={Vision language models are blind},
  author={Rahmanzadehgervi, Pooyan and Bolton, Logan and Taesiri, Mohammad Reza and Nguyen, Anh Totti},
  booktitle=ACCV,
  pages={18--34},
  year={2024}
}

@article{sun2026viprobe,
  title={Do {VLMs} Perceive or Recall? Probing Visual Perception vs. Memory with Classic Visual Illusions},
  author={Sun, Xiaoxiao and Li, Mingyang and Sun, Min Woo and Endo, Mark and Wu, Shengguang and Li, Changlin and Zhang, Yuhui and Wang, Zeyu and Yeung-Levy, Serena and others},
  journal={arXiv preprint arXiv:2601.22150},
  year={2026}
}

@article{hou2026viabench,
  title={Seeing Is Believing? A Benchmark for Multimodal Large Language Models on Visual Illusions and Anomalies},
  author={Hou, Wenjin and Liu, Wei and Hu, Han and Sun, Xiaoxiao and Yeung-Levy, Serena and Fan, Hehe},
  journal={arXiv preprint arXiv:2602.01816},
  year={2026}
}

@misc{gemini3flash_modelcard_2025,
  title  = {Gemini 3 Flash -- Model Card},
  author = {{Google DeepMind}},
  year   = {2025},
  month  = dec,
  note   = {Model card}
}

@misc{gemini31flashlite_modelcard_2026,
  title  = {Gemini 3.1 Flash-Lite -- Model Card},
  author = {{Google DeepMind}},
  year   = {2026},
  month  = mar,
  note   = {Model card}
}
}

\end{document}